\documentclass{article}

\usepackage{PRIMEarxiv}

\usepackage[utf8]{inputenc} 
\usepackage[T1]{fontenc}    
\usepackage{hyperref}       
\usepackage{url}            
\usepackage{booktabs}       
\usepackage{amsfonts}       
\usepackage{nicefrac}       
\usepackage{microtype}      
\usepackage{lipsum}
\usepackage{fancyhdr}       
\usepackage{graphicx}       
\graphicspath{{media/}}     
\usepackage{enumerate}
\usepackage{amsmath}
\usepackage{makecell}
\title{ArabianGPT: Native Arabic GPT-based Large Language Model} 

\author{
  Anis Koubaa, Adel Ammar, Lahouari Ghouti, Omar Najar, Serry Sibaee  \\
  Robotics and Internet-of-Things Lab \\
  Prince Sultan University \\
  Riyadh\\
  \texttt{\{akoubaa, aammar, lghouti, onajar, ssibaee\}@psu.edu.sa} \\
}

\begin{document}
\maketitle

\begin{abstract}
The predominance of English and Latin-based large language models (LLMs) has led to a notable deficit in native Arabic LLMs. This discrepancy is accentuated by the prevalent inclusion of English tokens in existing Arabic models, detracting from their efficacy in processing native Arabic's intricate morphology and syntax. Consequently, there is a theoretical and practical imperative for developing LLMs predominantly focused on Arabic linguistic elements. To address this gap, this paper proposes ArabianGPT, a series of transformer-based models within the ArabianLLM suite designed explicitly for Arabic. These models, including ArabianGPT-0.1B and ArabianGPT-0.3B, vary in size and complexity, aligning with the nuanced linguistic characteristics of Arabic. The AraNizer tokenizer, integral to these models, addresses the unique morphological aspects of Arabic script, ensuring more accurate text processing. Empirical results from fine-tuning the models on tasks like sentiment analysis and summarization demonstrate significant improvements. For sentiment analysis, the fine-tuned ArabianGPT-0.1B model achieved a remarkable accuracy of 95\%, a substantial increase from the base model's 56\%. Similarly, in summarization tasks, fine-tuned models showed enhanced F1 scores, indicating improved precision and recall in generating concise summaries. Comparative analysis of fine-tuned ArabianGPT models against their base versions across various benchmarks reveals nuanced differences in performance, with fine-tuning positively impacting specific tasks like question answering and summarization. These findings underscore the efficacy of fine-tuning in aligning ArabianGPT models more closely with specific NLP tasks, highlighting the potential of tailored transformer architectures in advancing Arabic NLP.
\end{abstract}




\maketitle

\section{Introduction}\label{sec1}

A notable theoretical and empirical challenge in natural language processing (NLP) is the underrepresentation of non-English languages in developing large language models (LLMs). This issue is particularly acute for languages with intricate linguistic structures, such as Arabic. The complexity of Arabic, with its rich morphology and diverse dialects, demands specialized attention in LLM development, a necessity often overlooked in the predominantly English-centric NLP landscape. This disparity limits the scope of NLP applications and hinders progress in understanding and processing multilingual contexts.
While GPT-1 and GPT-2 \cite{RAD2018, RAD2019} have laid the groundwork to demonstrate the capabilities of generative pre-trained transformers, their primary focus has been English, creating a gap in the Arabic NLP landscape. This gap becomes more evident when considering the success of models like Hulmona \cite{ELJ2019} and ARABERT \cite{ANT2020a}, which, despite their advances in Arabic NLP, still grapple with the challenges of adapting models trained predominantly on English data. 

Addressing this gap, ArabianGPT, within the broader ArabianLLM initiative, emerges as a targeted response. Unlike conventional LLMs, which are primarily designed to focus on English and exhibit a propensity to include a significant proportion of English tokens even in their multilingual models, ArabianGPT-Base is explicitly developed for the Arabic language. This model diverges from the norm by prioritizing Arabic's linguistic peculiarities in its design and implementation.

The methodology underpinning ArabianGPT-Base involves three critical components: an architectural adaptation, a rigorous training regimen, and a novel tokenization strategy. Firstly, ArabianGPT-Base adapts the proven GPT-2 architecture, tailoring it to accommodate Arabic's syntactic and morphological idiosyncrasies. This adaptation ensures that the model's structure is inherently aligned with the linguistic realities of Arabic, unlike conventional models that retrofit a predominantly English-centric architecture.

Secondly, the training of ArabianGPT-Base is conducted on a carefully curated corpus, specifically chosen to cover a wide range of Arabic text, thus enabling the model to learn from and adapt to the language's complexity and diversity. This aspect of the methodology is particularly critical, as it ensures that the model is exposed to and learns from a representative sample of the language rather than being limited to a subset that might bias its learning.

Lastly, ArabianGPT-Base introduces AraNizer, an innovative tokenizer developed to address the unique challenges of Arabic text segmentation specifically. This tokenizer represents a significant departure from standard tokenization methods, often inadequate for Arabic due to its complex script and morphological richness. AraNizer's design allows for more precise and contextually aware segmentation, a critical factor in enhancing the model's performance on downstream NLP tasks.

By integrating these methodological innovations, ArabianGPT-Base addresses the gap in Arabic NLP and sets a new standard for developing language-specific models. This approach marks a significant shift in the field, moving away from the one-size-fits-all strategy of previous LLMs towards a more nuanced and inclusive future for NLP.
\section{Related Works}\label{sec2}

\subsection*{Trends}


The evolution of Generative Pre-trained Transformers (GPTs) represents a significant milestone in Natural Language Processing (NLP), marking a departure from traditional models towards more sophisticated, contextually aware systems. Initiated by GPT-1 \cite{RAD2018}, the series showcased the efficacy of Causal Language Modeling (CLM) in understanding and generating human-like text. CLM, by predicting subsequent tokens based on previous context, enables deep learning models to capture nuanced linguistic patterns, facilitating both comprehension and production of complex language structures.

Progressing to GPT-3 \cite{BRO2020}, the field witnessed an exponential increase in model size and the diversification of training corpora. GPT-3, with its 175 billion parameters, leveraged an extensive dataset comprising vast swathes of internet text, setting new benchmarks across a multitude of NLP tasks. This leap in scale introduced advancements in few-shot learning, empowering models to perform tasks with minimal prior examples, a testament to their evolving contextual understanding.

Simultaneously, the advent of transfer learning emerged as a critical strategy for broadening the applicability of GPT models beyond English, addressing the challenges of linguistic diversity and dataset scarcity in languages like Arabic. The complexity of Arabic, with its rich morphological structure and dialectal variations, underscores the need for specialized approaches in data curation and model tuning.

This trajectory of GPT development not only underscores the importance of scale and data diversity in achieving linguistic comprehension but also highlights the ongoing challenges and opportunities in adapting these models for multilingual NLP, particularly for languages with unique syntactic and morphological features.

\subsection*{Existing Arabic LLMs}

\textbf{Bilingual Focus:} Models such as Jais \& Jais-Chat \cite{SEN2023} and AceGPT \cite{huang2023acegpt} have explored bilingual training regimes to enhance Arabic NLP capabilities. These approaches, while innovative, often struggle to fully capture the nuanced morphological and syntactic features of Arabic due to the inherent complexities of balancing bilingual corpora.

\textbf{Dialectal Emphasis:} Efforts by GigaBERT, ARBERT, and MARBERT \cite{lan2020empirical, abdul2020arbert} have been directed towards accommodating the rich dialectal diversity within the Arabic language. Despite their successes, these models may not fully address the intricacies of Modern Standard Arabic (MSA) and classical Arabic, limiting their applicability across all Arabic linguistic tasks.

\textbf{Understanding/Extraction Focus:} ARBERT, AraELECTRA, and CAMeLBERT \cite{ANT2020a, antoun2020araelectra, inoue2021interplay} have set new benchmarks in Arabic NLP for tasks centered around understanding and information extraction. However, their generative capabilities remain limited, underscoring a gap that ArabianGPT seeks to fill.

\textbf{Text-to-Text Focus:} Models like AraT5, QARiB, and JABER \& SABIR \cite{nagoudi2021arat5, abdelali2021pre, ghaddar2022revisiting} have demonstrated proficiency in text-to-text transformations. Nevertheless, their performance in generative tasks requiring deep linguistic comprehension of Arabic remains an area for further exploration.

\subsection*{AraNizer's Impact}

The AraNizer tokenizer stands as a testament to ArabianGPT's commitment to addressing the nuanced challenges of Arabic morphology. By intricately understanding and processing the unique morphological features of Arabic, AraNizer directly enhances ArabianGPT's ability to engage in more accurate and linguistically faithful text generation and processing tasks, setting it apart from the limitations noted in previous models.

\subsection*{ArabianGPT's Positioning}

ArabianGPT distinguishes itself by meticulously addressing the gaps left by existing Arabic LLMs. Through a deliberate design that emphasizes Arabic fluency and morphological understanding, ArabianGPT offers unparalleled capabilities in generative tasks. Its architecture, informed by the shortcomings of prior models, is specifically engineered to navigate the complexities of Arabic syntax and morphology, thereby facilitating advanced generative applications in native Arabic contexts, a leap forward in the pursuit of linguistic inclusivity and technological advancement in Arabic NLP.

ArabianGPT represents a significant stride towards addressing the multifaceted challenges of Arabic NLP. By harnessing specialized techniques such as the AraNizer tokenizer and focusing on the generative aspects of language processing, ArabianGPT not only bridges existing gaps but also paves the way for future innovations in the field.

\section{Datasets}\label{sec3}

\subsection{Dataset D1}\label{subsec2}

Dataset D1 represents a carefully curated compilation of contemporary Arabic newspaper articles, embodying an expansive collection of 5.37 million texts. This voluminous dataset, which aggregates to a substantial 15.5 GB, encompasses an extensive lexicon tallying 1.8 billion tokens, thereby providing a robust and comprehensive linguistic foundation for the ArabianGPT’s training regimen. This corpus encompasses a diverse array of news articles, each contributing to a broad spectrum of linguistic and cultural narratives spanning various chronological timelines. This diversity is paramount in ensuring that the model is exposed to a wide range of dialectical nuances, idiomatic expressions, and syntactic variations, all of which are intrinsic to modern Arabic vernacular and discourse.

The D1 Corpus is not simply a collection of texts; it is a systematic aggregation aimed at capturing the multifaceted nature of Arabic language and culture as represented in contemporary media. The corpus's vastness and variety provide a rich, contextual environment conducive to training sophisticated models capable of understanding and generating Arabic text with a high degree of linguistic fidelity. By covering a broad temporal range, from the early 2000s to 2014, the dataset facilitates ArabianGPT's comprehension of temporal linguistic shifts and trends, thereby enhancing its capability to produce pertinent and context-aware text.

This corpus has been strategically segmented and organized to facilitate efficient and effective training processes. Each article within the corpus has been pre-processed and structured to ensure optimal compatibility with the ArabianGPT's learning algorithms. The segmentation of the corpus into discrete articles allows for a granular approach to model training, where the nuances of each piece can be individually learned and integrated into the broader linguistic model.

The D1 Corpus, and consequently the ArabianGPT model, is limited by its knowledge cutoff in 2014. This restricts the model's understanding to the language, culture, and world events only up to that year, omitting subsequent developments and modern lexicon. This temporal limitation can impact the model's relevance and effectiveness in contemporary contexts, as it may produce outputs that are outdated or misaligned with recent trends and events. Users and developers must be aware of this constraint, especially for tasks requiring current knowledge.


\subsection{Dataset D2: Composite Linguistic Resource }\label{subsec2}

Dataset D2, employed for the training of ArabianGPT-0.3B, amalgamates three distinct yet complementary data sources. Collectively, these datasets constitute a comprehensive linguistic corpus with a substantial size of 23 GB and an impressive lexicon totaling approximately 3.34 billion (3,348,610,363) tokens. The strategic integration of these datasets is predicated on the premise of fostering a multifaceted and robust linguistic environment that is imperative for training a versatile and nuanced Arabic language model. 

Collectively, these datasets create a rich tapestry of linguistic scenarios, spanning from formal to colloquial and from historical to contemporary, thus equipping the ArabianGPT-0.3B model with the versatility needed to understand and generate Arabic text across a myriad of contexts and styles. It is this comprehensive and diverse linguistic environment that lays the groundwork for a robust and adaptable Arabic language model, capable of addressing a broad spectrum of language processing tasks with high efficacy.

\section{Tokenization}\label{sec4}

\subsection{Overview}
Effective natural language processing (NLP) hinges on accurate and efficient text tokenization, particularly for morphologically rich languages like Arabic. Standard tokenization methods often falter when confronted with the unique script, complex morphology, and extensive diacritics inherent in Arabic. This can lead to misinterpretations and hinder the performance of large language models (LLMs) trained on Arabic data.

To address these challenges, we introduced AraNizer, a custom-trained SentencePiece tokenizer specifically designed for Arabic NLP tasks. Leveraging the strengths of the SentencePiece model and incorporating domain-specific knowledge of Arabic linguistics, AraNizer delivers precise and semantically meaningful tokenization, paving the way for improved LLM performance.

Traditional tokenization techniques for Arabic often struggle with:

\begin{itemize}
    \item \textbf{Word root identification:} Separating prefixes, suffixes, and inflections from the core root morpheme poses challenges for standard tokenizers.
    
    \item \textbf{Diacritic handling:} Diacritics play a crucial role in Arabic morphology and semantics, yet their inconsistent representation and omission in some datasets can lead to ambiguities.
    
    \item \textbf{Ligature recognition:} The presence of ligatures, where multiple individual characters combine into a single glyph, further complicates word segmentation.
\end{itemize}

These limitations can adversely affect LLM performance on Arabic tasks, leading to inaccuracies in sentiment analysis, text classification, and other downstream applications. AraNizer aims to overcome these hurdles by:

\begin{itemize}
    \item \textbf{Fine-grained tokenization:} Utilizing the subword-level capabilities of SentencePiece, AraNizer achieves a deeper granularity than simple word segmentation, preserving morphological and semantic information within tokens.
    
    \item \textbf{Diacritic-aware processing:} AraNizer incorporates diacritics into its tokenization process, ensuring their contribution to word meaning is preserved and reflected in LLM representations.
    
    \item \textbf{Ligature handling:} AraNizer is trained to recognize and appropriately segment ligatures, avoiding misinterpretations and maintaining accurate morpheme boundaries.
\end{itemize}

\subsection{Description}

AraNizer \cite{ARA2023} is a custom-trained SentencePiece tokenizer \cite{KUD2018} specifically developed for the ArabianLMM series. It utilizes the SentencePiece model, empowering ArabianLMM to parse and generate coherent, contextually relevant Arabic text.
The Aranizer tokenizer addresses the complexities of Arabic text with remarkable effectiveness. It leverages SentencePiece and Byte Pair Encoding (BPE) \cite{shibata1999byte} methodologies to optimize tokenization accuracy and efficiency specifically for the Arabic language. This tailored approach ensures optimal handling of the language's morphological richness and unique script challenges.
To address the varying complexities and nuances of Arabic text, Aranizer is available in four distinct versions with vocabulary sizes of 32K, 50K, 64K, and 86K. This flexibility allows users to select the version that best suits their specific needs and application domains. For training ArabianGPT models, we opted for the 64K version, striking a balance between accuracy and efficiency.

\section{Small-scale model: ArabianGPT-0.1B}
\subsection{Model Overview}
The key objective of the unsupervised pre-training process is to train a language model that excels at predicting upcoming words in a text sequence, without requiring any labeled training data. The dataset D1, described in section 3.1, is used as the unsupervised corpus.
ArabianGPT model architecture employs a multi-layer transformer decoder, a powerful neural network architecture adept at capturing long-range dependencies within text. The Transformer architecture has proven highly effective in various language modeling tasks due to its ability to capture long-range dependencies.
The transformer decoder applies multi-headed self-attention to focus on relevant relationships between words within the context. The embedding layer transforms the input tokens U={} into numerical representations using token and position embeddings.
The objective function consists in maximizing the likelihood of predicting the next word in each sequence, given its preceding words. The objective function is expressed as:

\begin{equation}
    L(U) = \sum_{i} \log P(U_i | U_{i-k}, \ldots, U_{i-1}; \theta)
    \label{eq:likelihood}
\end{equation}

Where k is the context window size, and $\theta$ represents the model parameters (neural network’s weights) that are iteratively updated using stochastic gradient descent. 
In the ArabianGPT model architecture, the employed loss function characterizes the maximization of the conditional likelihood for sequential word prediction, thereby facilitating the model's proficiency in synthesizing contextually coherent and linguistically accurate Arabic text, a quintessential feature for advanced language understanding and generation. The trained model will ultimately generate a probability distribution over potential target tokens using a softmax function.
The model's ability to predict upcoming words stems from its learned understanding of language patterns and relationships between words. Pre-training on a massive, diverse corpus is crucial for the model to capture a broad range of linguistic phenomena.

\subsection{Description}
\begin{figure}[h]
  \centering
  \includegraphics[width=7cm]{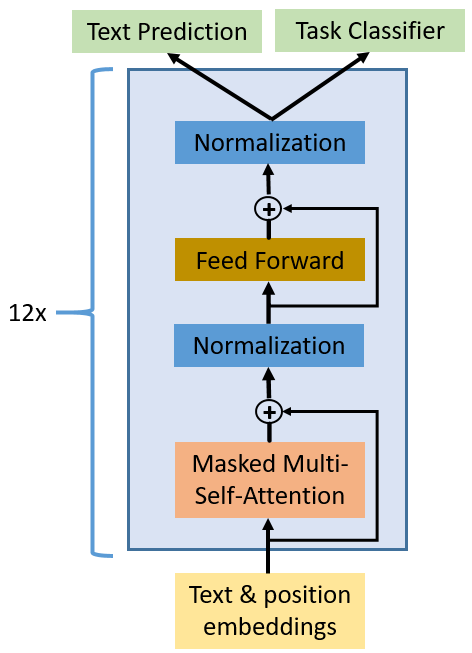}
  \caption{Architecture of ArabianGPT-0.1 model based on GPT2-small.}
  \label{fig:Architecture-model-0.1}
\end{figure}

ArabianGPT-0.1 follows the architecture of GPT-2-small \cite{RAD2019}, which is basically equivalent to GPT1 \cite{RAD2018} (Figure \ref{fig:Architecture-model-0.1}). We retain the underlying principles of the GPT model while incorporating modifications (especially the tokenizer) to better suit Arabic text processing.
The ArabianGPT-0.1 model represents a significant advancement in the field of Natural Language Processing (NLP) for Arabic language. Leveraging the GPT-2 architecture, ArabianGPT is custom-tailored to grasp and generate pure Arabic text.  
Figure \ref{fig:Architecture-model-0.1} shows the main components of this GPT model. It consists of an embedding layer followed by 12 identical layers, each made up of three sub-layers: masked multi-head self-attention, a feed-forward network, and layer normalization. 
This stacked architecture allows the model to learn complex relationships between words in a sentence:

\begin{itemize}
    \item \textbf{Text and Position Embedding Layer:} Adds information about the individual words and their positions in the sentence, which helps the model to better understand the meaning of the text.
    
    \item \textbf{Masked Multi-Head Self-Attention:} The core component of the GPT-1 architecture. It allows the model to attend to different parts of the input sentence simultaneously, using multiple heads that focus on different aspects of the words. This gives the model the ability to capture long-range dependencies and context in the text.
    
    \item \textbf{Feed-Forward Sub-Layer:} Adds non-linearity to the model, allowing it to learn more complex relationships between words.
    
    \item \textbf{Normalization Layer:} Helps to stabilize the training process and improve the model's overall performance.
\end{itemize}

The ArabianGPT model inherits several key features from the GPT-2 model:

\begin{itemize}
    \item \textbf{Unidirectional Language Modeling:} ArabianGPT, like GPT-2, predicts the probability of a sequence of words in a text, learning to generate coherent and contextually relevant text based on the input it receives.
    
    \item \textbf{Transformer-based Architecture \cite{VAS2017}:} The use of 12 Transformer layers aids in capturing long-range dependencies in text, a crucial aspect for Arabic text which often involves complex sentence structures.
    
    \item \textbf{Attention Mechanism:} Each Model Attention Layer in ArabianGPT is designed to focus on different parts of the input text, allowing the model to generate more nuanced and accurate Arabic text.
    
    \item \textbf{Adaptive Learning Rate:} ArabianGPT leverages an adaptive learning rate during training, which facilitates efficient and effective learning from the large-scale Arabic corpus.
\end{itemize}

\subsection{Specifications}

Table \ref{tab:Hyperparameter-ArabianGPT-0.1} shows the main characteristics and hyperparameters of the ArabianGPT-0.1 model. 

\begin{table}[htbp]
  \centering
  \caption{Characteristics and hyperparameters of the ArabianGPT-0.1 model.}\label{tab:Hyperparameter-ArabianGPT-0.1}
  \begin{tabular}{ll}
    \toprule
    \textbf{Attribute} & \textbf{Value} \\
    \midrule
    Model Name & ArabianGPT-0.1 \\
    Architecture & GPT-2 (GPT2LMHeadModel) \\
    Number of Layers & 12 \\
    Number of Model Attention Layers (MALs) & 12 \\
    Number of Heads & 12 \\
    Vocabulary Size & 64,000 \\
    Tokenizer & Aranizer \\
    Model Size & 134M parameters \\
    Initial Learning Rate & $4 \times 10^{-5}$ \\
    Final Learning Rate & $4 \times 10^{-0.4}$ \\
    Context Window Size & 768 tokens \\
    Dropout Ratio for Attention (attn\_pdrop) & 0.1 \\
    Dropout Ratio for Embeddings (embd\_pdrop) & 0.1 \\
    Dropout Probability for all Fully Connected Layers (resid\_pdrop) & 0.1 \\
    \bottomrule
  \end{tabular}
\end{table}

\subsection{Training procedure}

\begin{figure}[h]
  \centering
  \includegraphics[width=14cm]{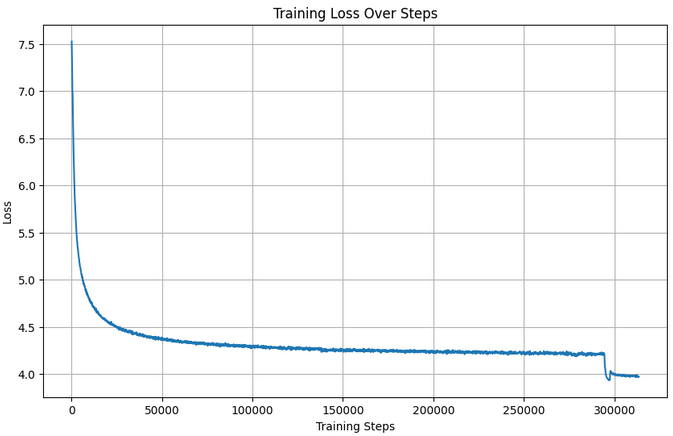}
  \caption{Evolution of the training loss for the ArabianGPT-0.1 model.}
  \label{fig:loss-model-0.1}
\end{figure}

ArabianGPT-0.1B was trained on dataset D1 (see section 3.1) using 2 NVIDIA A100 GPUs with 80GB each. The A100 GPUs are known for their efficiency and capability in handling large-scale deep learning tasks. The model underwent training with a dataset comprising 7.5 million sequences, each with a sequence length of 768 tokens. The training parameters included a batch size of 512, a total of 313,500 steps, and a duration of 3 days, resulting in a final loss of 3.97. The training was focused on adapting the model to recognize and interpret the nuances and complexities inherent in Arabic. Table \ref{tab:training-ArabianGPT-0.1} summarizes the conditions and parameters of the training, and Figure \ref{fig:loss-model-0.1} shows the training loss convergence through the steps of the training.

\begin{table}[htbp]
  \centering
  \caption{Details of the training of ArabianGPT-0.1B.}\label{tab:training-ArabianGPT-0.1}
  \begin{tabular}{ll}
    \toprule
    \textbf{Parameter} & \textbf{Value} \\
    \midrule
    Model & ArabianGPT-0.1B (Base Model) \\
    Hardware & 2 NVIDIA A100 GPU (80GB each) \\
    Number of Examples & 7.5 million sequences (sequence length = 768 tokens) \\
    Batch Size & 512 \\
    Number of Steps & 313,500 \\
    Number of Epochs & 6 \\
    Training Duration & 3 days \\
    Final Loss & 3.97 \\
    \bottomrule
  \end{tabular}
\end{table}

\subsection{Testing Examples}

Table \ref{tab:sample-output-ArabianGPT-0.1} shows some testing prompt and completion pairs for the ArabianGPT-0.1B model. We notice that the completions seem natural, with a few punctuation issues.

\begin{table}[h]
  \centering
  \caption{Sample output of the ArabianGPT-0.1B model.}\label{tab:sample-output-ArabianGPT-0.1}
  \includegraphics[width=12cm]{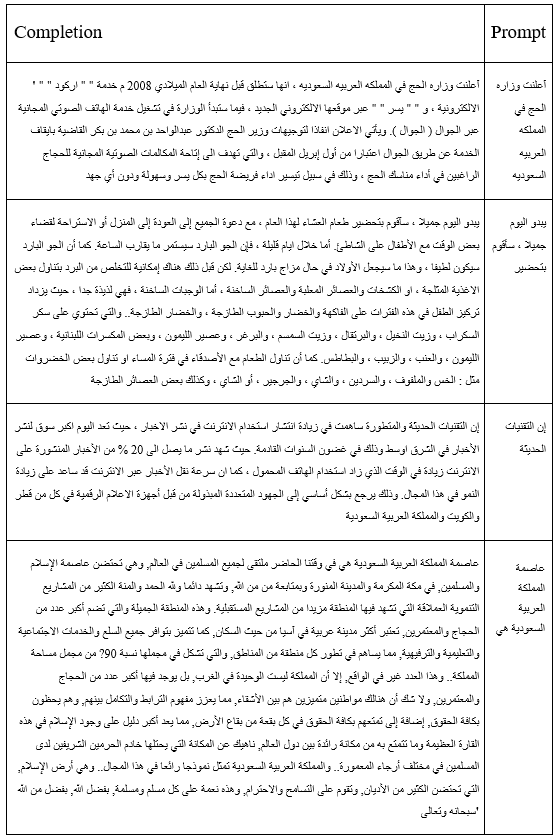}
\end{table}

\section{Medium-scale model: ArabianGPT-0.3B}

\subsection{Model Overview}

Compared to its predecessor, ArabianGPT-0.1B, ArabianGPT-0.3B is significantly larger and more advanced, trained on a vastly increased dataset of examples. This allows it to capture the intricacies and nuances of Arabic across various domains and topics with greater accuracy and effectiveness.

\subsection{Description}

ArabianGPT-0.3B is a variant of the GPT-2 model \cite{RAD2019} with a medium size, specifically pre-trained for the Arabic language. It retains the underlying principles of GPT-2 while incorporating modifications to better suit Arabic text processing. ArabianGPT-0.3B is a more advanced and larger version than ArabianGPT-0.1. It handles a significantly larger number of examples, making it more adept at capturing the nuances and complexities of the Arabic language across various topics.

\subsection{Specifications}
Table \ref{tab:hyperparameters-model-0.5} shows the main characteristics and hyperparameters of the ArabianGPT-0.3 model. It has a double number of layers compared to ArabianGPT-0.1, with a larger number of Model Attention Layers (15 compared to 12) and heads (15 compared to 12).

\begin{table}[h]
  \centering
  \caption{Characteristics and hyperparameters of the ArabianGPT-0.5 model.}\label{tab:hyperparameters-model-0.5}
  \begin{tabular}{ll}
    \toprule
    \textbf{Attribute} & \textbf{Value} \\
    \midrule
    Model Name & ArabianGPT-MID \\
    Architecture & GPT-2 (GPT2LMHeadModel) \\
    Number of Layers & 24 \\
    Number of Model Attention Layers (MALs) & 16 \\
    Number of Heads & 16 \\
    Vocabulary Size & 64,000 \\
    Model Size & 345M parameters \\
    Initial Learning Rate & $4 \times 10^{-5}$ \\
    Final Learning Rate & $1.1 \times 10^{-0.4}$ \\
    Context Window Size & 1024 tokens \\
    Dropout Ratio for Attention (attn\_pdrop) & 0.1 \\
    Dropout Ratio for Embeddings (embd\_pdrop) & 0.1 \\
    Dropout Probability for all Fully Connected Layers (resid\_pdrop) & 0.1 \\
    \bottomrule
  \end{tabular}
\end{table}

\subsection{Training Procedure}
\begin{figure}[h]
  \centering
  \includegraphics[width=14cm]{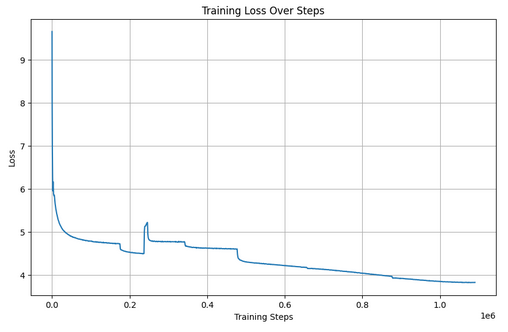}
  \caption{Evolution of the training loss for the ArabianGPT-0.3 model.}
  \label{fig:loss-model-0.3}
\end{figure}

Table \ref{tab:training_model-0.3} summarizes the conditions and parameters of the training of the ArabianGPT-0.3B model. Utilizing the power of 4 NVIDIA A100 GPUs, the model underwent an extensive training process, leveraging a dataset consisting of 70 million sequences with a sequence length of 1024. With a batch size of 512 and 1,009,500 steps over 4.97 epochs, the training duration spanned 4.1 days. The culmination of this training journey resulted in a final loss of 3.82, as illustrated in Figure \ref{fig:loss-model-0.3}, depicting the convergence of the training loss throughout the steps of the training process.
 
\begin{table}[h] 
  \centering
  \caption{Details of the training of ArabianGPT-0.3B.}\label{tab:training_model-0.3}
  \begin{tabular}{ll}
    \toprule
    \textbf{Parameter} & \textbf{Value} \\
    \midrule
    Model & ArabianGPT-0.3B (Medium Model) \\
    Hardware & 4 NVIDIA A100 GPU (80GB each) \\
    Number of Examples & 70 million sequences (sequence length = 1024) \\
    Batch Size & 512 \\
    Number of Steps & 1,009,500 \\
    Number of Epochs & 4.97 \\
    Training Duration & 4.1 days \\
    Final Loss & 3.82 \\
    \bottomrule
  \end{tabular}
\end{table}

\subsection{Testing examples}

Table \ref{tab:sample-output-ArabianGPT-0.3} shows some testing prompt and completion pairs for the ArabianGPT-0.1B model. We notice that the completions seem natural, with better punctuation.

\begin{table}[h]
  \centering
  \caption{Sample output of the ArabianGPT-0.3B model.}\label{tab:sample-output-ArabianGPT-0.3}
  \includegraphics[width=15cm]{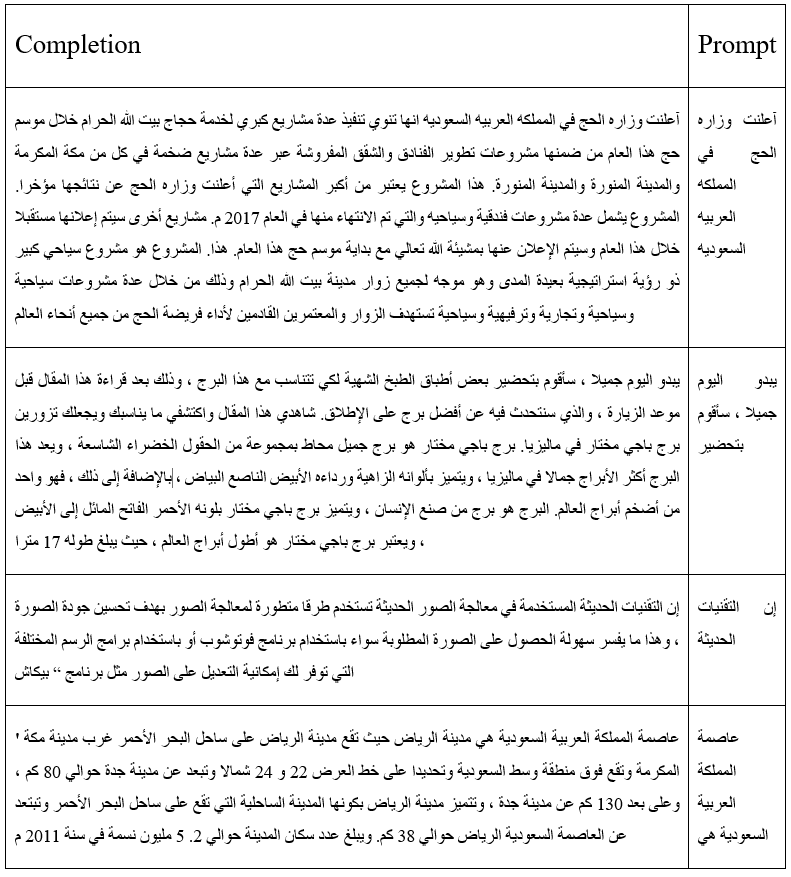}
\end{table}

\section{Few-shot evaluation}\label{sec:few-shot-eval}
In this section, we present a detailed analysis of the performance of ArabianGPT-0.1B, ArabianGPT-0.3B, and compare their performance to AraGPT-Base, AraGPT-Medium, Bloom-7b1, and Llama-7B models across various benchmarks. This evaluation follows a few-shot language model evaluation framework, providing insights into the models' abilities in scientific reasoning, common-sense understanding, cross-domain knowledge, and truthfulness. The benchmarks are designed to test different aspects of language models. They include:

\begin{itemize}
    \item \textbf{ARC (25-shot) normalized:} Assesses scientific reasoning capabilities using normalized metrics.
    \item \textbf{HellaSwag (10-shot):} Evaluates common-sense reasoning in narrative contexts.
    \item \textbf{MMLU (5-shot):} Measures understanding across various subjects.
    \item \textbf{TruthfulQA (0-shot) with mc2:} Gauges the ability to provide truthful and accurate responses.
\end{itemize}

Table \ref{tab:few_shot_eval} displays the evaluation scores of the aforementioned models on each benchmark.  ArabianGPT-0.3B shows a robust aggregate performance across all tasks. It notably excels in TruthfulQA (zero-shot) with a score of 52.5, exceeding Bloom-7b1, and Llama-7B models, which are more than 23 times larger. ArabianGPT-0.1B closely trails its larger counterpart. Despite its smaller size, it competes closely with ArabianGPT-0.3B, particularly in ARC (25-shot) and MMLU (5-shot). The performance efficiency of ArabianGPT-0.1B is notable, given its fewer parameters, indicating effective learning and adaptation capabilities. This demonstrates that smaller models can still deliver competitive performance in diverse linguistic tasks.

\begin{table}[h]
    \centering
    \caption{Zero and Few-shot Evaluation Scores}\label{tab:few_shot_eval}
    \begin{tabular}{lcccccc}
        \toprule
\textbf{Model}              & \textbf{Language} & \textbf{Average} & \textbf{\begin{tabular}[c]{@{}c@{}}ARC \\ (25-shot)\end{tabular}} & \textbf{\begin{tabular}[c]{@{}c@{}}HellaSwag \\ (10-shot)\end{tabular}} & \textbf{\begin{tabular}[c]{@{}c@{}}MMLU \\ (5-shot)\end{tabular}} & \textbf{\begin{tabular}[c]{@{}c@{}}TruthfulQA \\ (0-shot)\end{tabular}} \\ \hline
        \midrule
        Bloom-7b1 & Multilingual & 36.2 & 31.4 & 43.3 & 27.5 & 42.6 \\
        Llama-7B & Multilingual & 32.1 & 24.6 & 30.9 & 28.0 & 45.1 \\
        ArabianGPT-0.3B & Arabic & 32.7 & 24.3 & 28.4 & 25.7 & 52.5 \\
        ArabianGPT-0.1B & Arabic & 31.9 & 24.0 & 26.6 & 25.4 & 51.8 \\
        AraGPT-Base & Arabic & 31.7 & 24.6 & 27.5 & 25.1 & 49.5 \\
        AraGPT-Medium & Arabic & 32.2 & 23.9 & 28.5 & 26.3 & 50.0 \\
        \bottomrule
    \end{tabular}
\end{table}

\section{Fine Tuning}
\subsection{Summarization}
The task of summarization in natural language processing (NLP) involves generating concise and meaningful summaries from a larger text. In this project, we have fine-tuned the previously pre-trained ArabianGPT 0.1B and 0.3B Models for the summarization task. These models have been fine-tuned (FT) on a summarization dataset, with the aim of improving their ability to generate accurate and relevant summaries.
The performance of these models has been evaluated using the F1 score, a metric that combines precision and recall.

The dataset used for fine-tuning and evaluation is a subset of xlsum \cite{xlsum_arabic}, specifically xlsum[1000]. This dataset comprises a range of texts suitable for testing summarization capabilities.

The fine-tuning process involves adjusting the weights of the pretrained model specifically for the summarization task. This phase is crucial as it tailors the model to perform better on the desired task, here summarization.

\subsubsection{Hyperparameters}
In the fine-tuning process of the ArabianGPT models, the Cross-Entropy Loss function was employed alongside the ADAM optimizer. The Cross-Entropy Loss function is widely used in classification tasks, including tasks like summarization and question answering, which can be formulated as predicting the next token in a sequence.

\textbf{Cross-Entropy Loss:} This loss function measures the performance of a classification model whose output is a probability value between 0 and 1. It increases as the predicted probability diverges from the actual label, making it suitable for training models in tasks like summarization and question answering.

\textbf{Optimizer:} ADAM, an algorithm for first-order gradient-based optimization, is utilized for its efficiency in handling sparse gradients and its adaptability to large-scale problems. It's a popular choice in fine-tuning language models due to its effective handling of large datasets and complex neural network architectures.

\begin{figure}[h]
  \centering
  \includegraphics[width=17cm]{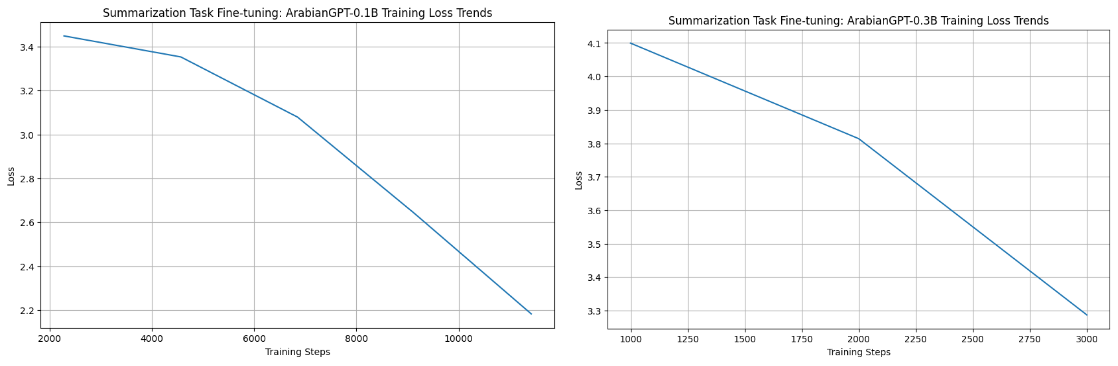}
  \caption{Fine-tuning loss for the ArabianGPT-0.1B and ArabianGPT-0.3B models.}
  \label{fig:loss-summarization}
\end{figure}

Figure \ref{fig:loss-summarization} depicts the fine-tuning loss for the ArabianGPT-0.1B and ArabianGPT-0.3B models. The smooth decrease in loss values from initial to final stages indicates successful learning and adaptation during fine-tuning.
The initial loss values provide a benchmark to gauge the effectiveness of the fine-tuning process. The significant reduction in these values by the end of fine-tuning reflects the models' improved ability to predict accurate tokens for the tasks at hand.
The difference in loss values between the Summarization and Question Answering tasks suggests that these tasks have varying levels of complexity or that the models respond differently to the intricacies of each task.

The ArabianGPT0.3B models start with a higher initial loss in both tasks compared to the 0.1B models, potentially indicating a more complex model capacity which, after fine-tuning, also shows a substantial reduction in loss.

\subsubsection{Results}
The performance of both the original and fine-tuned models was evaluated using the F1 score. The F1 score is a harmonic mean of precision and recall, providing a balance between the completeness and accuracy of the summaries generated by the models. They are calculated using BLUE and ROUGE Scores using the following equation:
\[ F1 = \frac{2 \cdot \text{BLUE} \cdot \text{ROUGE}}{\text{BLUE} + \text{ROUGE}} \]

\begin{figure}[h]
  \centering
  \includegraphics[width=14cm]{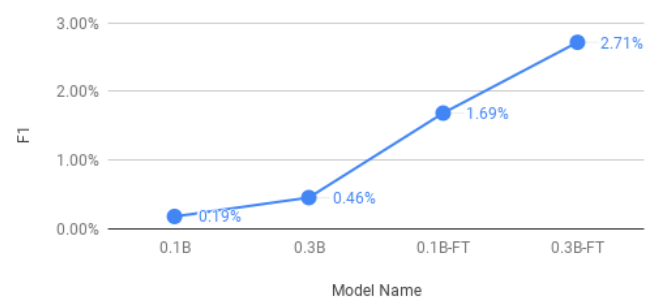}
  \caption{F1 score for the ArabianGPT-0.1B and ArabianGPT-0.3B models, before and after fine-tuning on the Summarization dataset.}
  \label{fig:F1-summarization}
\end{figure}

Figure \ref{fig:F1-summarization} depicts the F1 score for the ArabianGPT-0.1B and ArabianGPT-0.3B models, before and after fine-tuning on the Summarization dataset. The fine-tuned models (0.1B-FT and 0.3B-FT) demonstrate superior performance compared to their respective non-fine-tuned counterparts. The observed increase in F1 scores post fine-tuning suggests that the models have enhanced their proficiency in summarization tasks. Notably, the 0.3B models, both fine-tuned and non-fine-tuned, exhibit better performance than their 0.1B counterparts, indicating that a larger model capacity can effectively handle the complexity of summarization tasks. Despite these advancements, the F1 scores remain relatively low, signifying substantial room for improvement. Possible avenues for addressing this include further fine-tuning, utilizing a larger and more diverse dataset, or experimenting with different model architectures and hyperparameters.

Tables \ref{tab:sample-output-ArabianGPT-0.1} and \ref{tab:sample-output-ArabianGPT-0.3} show a sample output of the fine-tuned ArabianGPT-0.1B and ArabianGPT-0.3B models, respectively, on the summarization task. We notice that the models understand the task and summarize the main idea, but the output sometimes include additional irrelevant words.

Table \ref{tab:eval-summary} shows the scores obtained by the fine-tuned ArabianGPT-0.1B and ArabianGPT-0.3B models on the benchmarks introduced in section \ref{sec:few-shot-eval}. We notice that the two models have close performance in average. But ArabianGPT-0.3B is significantly better on ARC and HellaSwag benchmarks, while it lags markedly behind ArabianGPT-0.1B on the TruthfulQA benchmark.

\begin{table}[h]
  \centering
  \caption{Sample output of the fine-tuned ArabianGPT-0.1B model on the summarization task.}\label{tab:sample-output-ArabianGPT-0.1}
  \includegraphics[width=15cm]{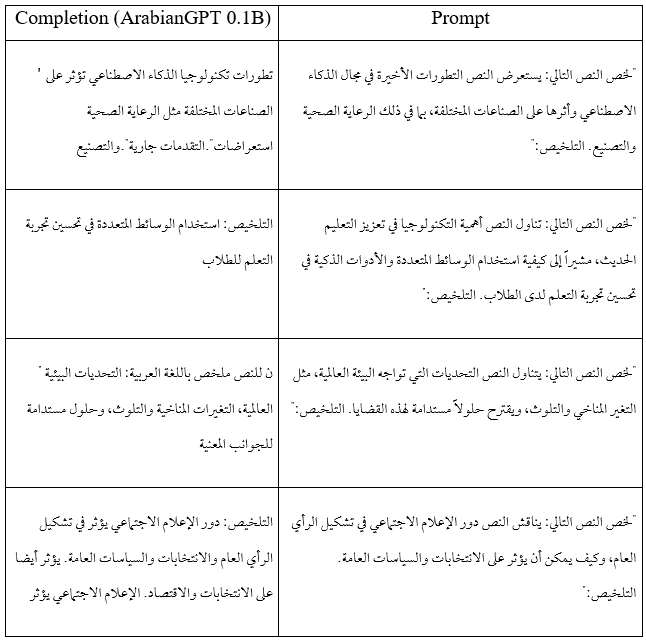}
\end{table}

\begin{table}[h]
  \centering
  \caption{Sample output of the fine-tuned ArabianGPT-0.3B model on the summarization task.}\label{tab:sample-output-ArabianGPT-0.3}
  \includegraphics[width=15cm]{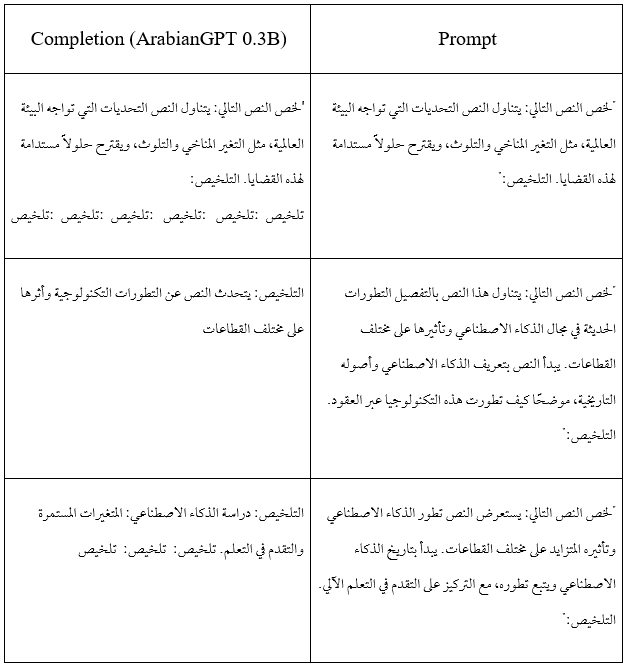}
\end{table}

\begin{table}[htbp]
    \centering
    \caption{Model Evaluation Scores for ArabianGPT-0.1B and ArabianGPT-0.3B after fine tuning on the Summarization dataset}\label{tab:eval-summary}
    \begin{tabular}{lcccccc}
        \toprule
\textbf{Model}              & \textbf{Language} & \textbf{Average} & \textbf{\begin{tabular}[c]{@{}c@{}}ARC \\ (25-shot)\end{tabular}} & \textbf{\begin{tabular}[c]{@{}c@{}}HellaSwag \\ (10-shot)\end{tabular}} & \textbf{\begin{tabular}[c]{@{}c@{}}MMLU \\ (5-shot)\end{tabular}} & \textbf{\begin{tabular}[c]{@{}c@{}}TruthfulQA \\ (0-shot)\end{tabular}} \\ \hline
\textbf{ArabianGPT-0.1B-FT} & Arabic            & 31.7             & 24.5                                                              & 28.0                                                                    & 25.0                                                              & 49.3                                                                    \\
\textbf{ArabianGPT-0.3B-FT} & Arabic            & 31.8             & 27.3                                                              & 31.5                                                                    & 25.1                                                              & 43.2                                                                    \\ \hline
\end{tabular}
\end{table}
The fine-tuning of ArabianGPT models on the summarization task has demonstrated improvements in their performance as indicated by the F1 scores. The results suggest a positive correlation between model size and summarization ability, as well as the effectiveness of fine-tuning in enhancing model performance for specific tasks. Future work could focus on exploring advanced fine-tuning techniques, larger and more diverse datasets, and possibly integrating other AI methodologies to further enhance the summarization capabilities of transformer-based models.

\subsection{Sentiment Analysis}
We fine-tuned the  Arabic Jordanian General Tweets (AJGT) Corpus, which consists of 1,800 tweets annotated as positive and negative. They are expressed either in Modern Standard Arabic (MSA) or Jordanian dialect. Table \ref{tab:sent_analysis_training_specs} details the specifications for the sentiment analysis fine-tuning task.

Table \ref{tab:sent_analysis_perf} compares the performance of the ArabianGPT-0.1B model on the sentiment analysis task, before and after fine tuning. While the base model gave almost random responses, we notice a marked increase in accuracy after fine tuning attaining 95\% after only 3 epochs, on a relatively small fine-tuning dataset.

\begin{figure}[h]
  \centering
  \includegraphics[width=14cm]{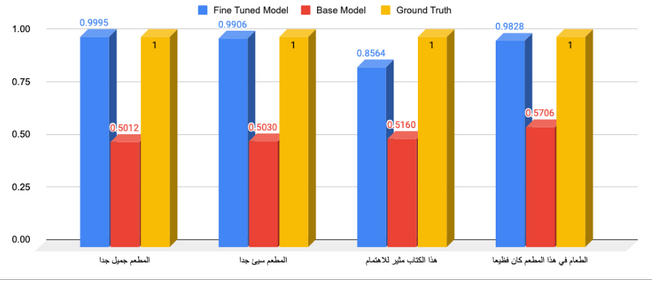}
  \caption{Comparison of the output score of the ArabianGPT-0.1B model before and after fine tuning on the sentiment analysis task, on sample inputs.}
  \label{fig:sample_sentim_analysis}
\end{figure}

Table \ref{tab:sample-output-ArabianGPT-0.1-sentim-analysis} and Figure \ref{fig:sample_sentim_analysis} show a sample output of the ArabianGPT-0.1B model before and after fine tuning on the sentiment analysis task. Before fine tuning, the output scores are close to 0.5, which indicates that the model responses are almost random. By contrast, after fine tuning, the scores are close to 1, and the output is correct in most cases.

\begin{table}[h]
    \centering
    \caption{Training Specifications for the sentiment analysis fine-tuning task.}\label{tab:sent_analysis_training_specs}
    \begin{tabular}{ll}
        \hline
        \textbf{Parameter} & \textbf{Value} \\
        \hline
        Epochs & 3 \\
        Learning rate & $1.4 \times 10^{-5}$ to $1.7 \times 10^{-6}$ \\
        Batch & 8 \\
        GPU & A100 \\
        Global Step & 1500 \\
        Train Loss & 0.203 \\
        Training Samples & 375 \\
        Testing Samples & 375 \\
        Training Percentage & 70\% \\
        Testing Percentage & 30\% \\
        Tokenizer & 64K aranizer \\
        \hline
    \end{tabular}
\end{table}

\begin{table}[h]
    \centering
    \caption{Model Performance on the sentiment analysis task, before and after fine tuning.}\label{tab:sent_analysis_perf}
    \begin{tabular}{l c}
        \hline
        \textbf{Model} & \textbf{Accuracy} \\
        \hline
        LLM-0.1B-Base & 56\% \\
        LLM-0.1B-Base (fine-tuned) & 95\% \\
        \hline
    \end{tabular}
\end{table}

\begin{table}[h]
  \centering
  \caption{Sample output of the ArabianGPT-0.1B model before and after fine tuning on the sentiment analysis task.}\label{tab:sample-output-ArabianGPT-0.1-sentim-analysis}
  \includegraphics[width=15cm]{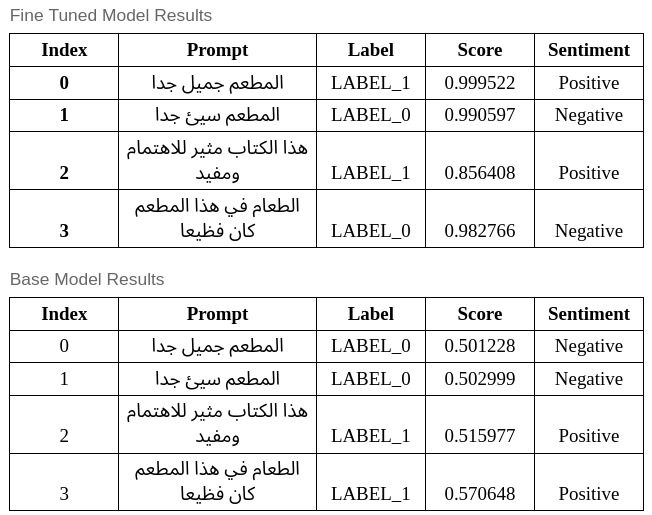}
\end{table}

\subsection{Question Answering}
The task of question-answering in natural language processing (NLP) involves generating accurate and contextually relevant answers to given queries. We have fine-tuned the previously pre-trained ArabianGPT 0.1B and 0.3B Models for the question-answering task. These models have undergone fine-tuning (FT) on a question-answering dataset, specifically the mkqa-1000 dataset, with the goal of enhancing their capability to provide precise and informative answers.

The mkqa-1000 dataset, derived from the Google Natural Questions dataset, serves as the basis for fine-tuning and evaluation. This dataset encompasses a diverse set of queries and passage-independent answers, which facilitates testing the models' question-answering capabilities across various contexts.

The fine-tuning process involves adjusting the weights of the pretrained model, tailoring it specifically for the question-answering task. This crucial phase aims to optimize the model's performance and enhance its proficiency in generating accurate responses to a wide range of queries in Arabic language.

Table \ref{tab:eval-ft-qa} shows the scores obtained by the fine-tuned ArabianGPT-0.1B and ArabianGPT-0.3B models on the benchmarks introduced in section \ref{sec:few-shot-eval}. We notice that the two models have close performance in average. But ArabianGPT-0.3B is significantly better on HellaSwag benchmarks, while it lags behind ArabianGPT-0.1B on the TruthfulQA benchmark, suggesting that it was not trained enough with respect to its larger number of parameters.

Table \ref{tab:sample_output_qa} displays a sample output of our two models after fine tuning on the Question/Answering dataset. While ArabianGPT-0.1B commonly yields irrelevant, incomplete, or unclear responses, some of ArabianGPT-0.3B responses are perfectly accurate, such as for the first and third example.

\begin{table}[htbp]
    \centering
    \caption{Model Evaluation Scores for ArabianGPT-0.1B and ArabianGPT-0.3B after fine tuning on the Question/Answering dataset}\label{tab:eval-ft-qa}
    \begin{tabular}{lcccccc}
        \toprule
        \textbf{Model} & \textbf{Language} & \textbf{Average} & \makecell{\textbf{ARC} \\ \textbf{(25-shot)}} & \makecell{\textbf{HellaSwag} \\ \textbf{(10-shot)}} & \makecell{\textbf{MMLU} \\ \textbf{(5-shot)}} & \makecell{\textbf{TruthfulQA} \\ \textbf{(0-shot)}} \\
        \midrule
        ArabianGPT-0.1B-FT & Arabic & 31.2 & 25.2 & 25.0 & 24.9 & 49.6 \\
        ArabianGPT-0.3B-FT & Arabic & 31.4 & 24.8 & 27.6 & 24.9 & 48.2 \\
        \bottomrule
    \end{tabular}
\end{table}

\begin{table}[h]
  \centering
  \includegraphics[width=14cm]{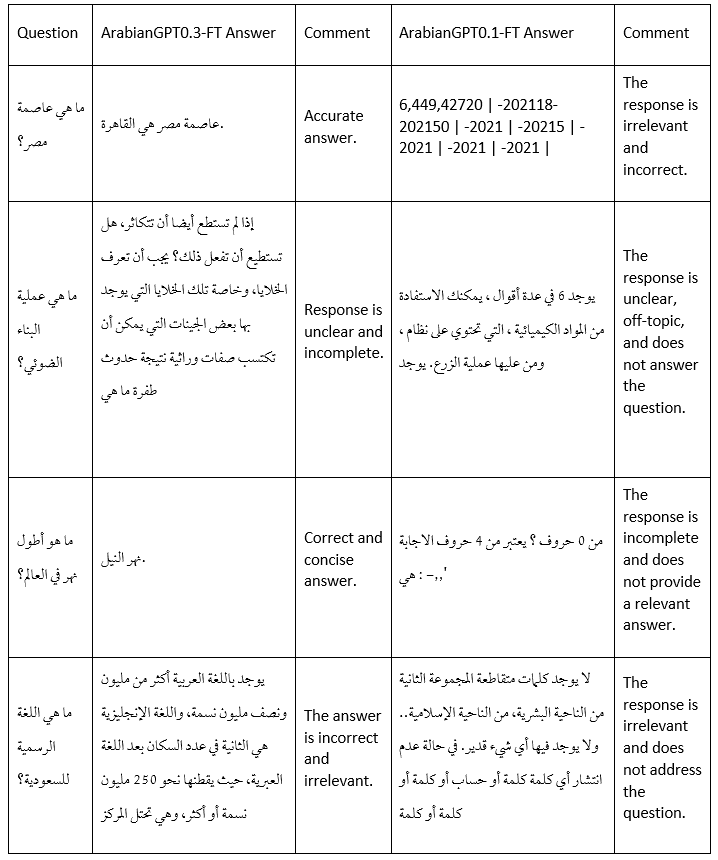}
  \caption{Sample output of the two ArabianGPT Models after fine tuning on the Question/Answering dataset.}
  \label{tab:sample_output_qa}
\end{table}
 
\section{Conclusion}\label{sec13}
The ArabianGPT-Base models represent a significant leap forward in the realm of Arabic natural language processing (NLP), introducing robust capabilities for both understanding and generating Arabic text. The architecture's strength, rooted in its training on a comprehensive corpus and the utilization of the customized tokenizer, AraNizer, collectively contribute to its effectiveness across various NLP tasks specific to the Arabic language context. Notably, the versatility demonstrated by the GPT-2 architecture is further emphasized by the successful adaptation to the intricacies of the Arabic language through the ArabianGPT models.

The development of ArabianGPT is driven by the acknowledgment of a substantial gap in the availability of native Arabic large language models (LLMs), primarily due to the prevalence of English and Latin-based models. Existing models often suffer from the inclusion of English tokens, which compromises their efficacy in processing the intricate morphology and syntax inherent to native Arabic. This paper introduced ArabianGPT as a tailored solution within the ArabianLLM suite, explicitly designed to address this deficit. The suite includes models of varying sizes, such as ArabianGPT-0.1B and ArabianGPT-0.3B, aligning with the nuanced linguistic characteristics of Arabic.

A key component of the ArabianGPT models is the AraNizer tokenizer, specifically crafted to address the unique morphological aspects of Arabic script. This ensures more accurate text processing, allowing the models to navigate the intricacies of the language with precision. The empirical results from fine-tuning these models on tasks like sentiment analysis and summarization showcase significant improvements. For instance, the fine-tuned ArabianGPT-0.1B model achieved an impressive accuracy of 95\%, a substantial increase from the base model's 56\%, in sentiment analysis. Similarly, in summarization tasks, fine-tuned models exhibited enhanced F1 scores, indicating improved precision and recall in generating concise summaries.

Comparative analyses of the fine-tuned ArabianGPT models against their base versions across various benchmarks reveal nuanced differences in performance. Fine-tuning is observed to have a positive impact on specific tasks, such as question answering and summarization. These findings underscore the efficacy of fine-tuning in aligning ArabianGPT models more closely with specific NLP tasks, highlighting the potential of tailored transformer architectures in advancing Arabic NLP. As such, the ArabianGPT-Base models not only contribute to bridging the existing gap in Arabic LLMs but also pave the way for further advancements in the field of Arabic NLP.

\clearpage

\section*{Funding}
This research was funded by Prince Sultan University, grant number SEED-CCIS-2023-145.

\section*{Institutional Review Board Statement}
Not applicable.

\section*{Informed Consent Statement}
Not applicable.

\section*{Data Availability Statement}
The datasets used in this work are publicly available on https://huggingface.co/datasets.

\section*{Acknowledgments}
The authors thank Prince Sultan University for supporting this work.

\bibliography{sn-bibliography}

\bibliographystyle{unsrt} 
\end{document}